\documentclass[letterpaper]{article} % DO NOT CHANGE THIS
\usepackage{aaai2026}

\usepackage{times}  % DO NOT CHANGE THIS
\usepackage{helvet}  % DO NOT CHANGE THIS
\usepackage{courier}  % DO NOT CHANGE THIS
\usepackage[hyphens]{url}  % DO NOT CHANGE THIS
\usepackage{graphicx} % DO NOT CHANGE THIS
\usepackage{natbib}  % DO NOT CHANGE THIS AND DO NOT ADD ANY OPTIONS TO IT
\usepackage{caption} % DO NOT CHANGE THIS AND DO NOT ADD ANY OPTIONS TO IT
\usepackage{algorithm}
\usepackage{algorithmic}

\usepackage{newfloat}
\usepackage{listings}
\DeclareCaptionStyle{ruled}{labelfont=normalfont,labelsep=colon,strut=off} % DO NOT CHANGE THIS
\floatstyle{ruled}
\newfloat{listing}{tb}{lst}{}
\floatname{listing}{Listing}

\usepackage{booktabs}
\usepackage{amsmath}
\usepackage{amssymb}
\usepackage{multirow}

\title{Auto-PRE: An Automatic and Cost-Efficient Peer-Review Framework for Language Generation Evaluation}
\author{
    Junjie Chen\textsuperscript{\rm 1},
    Weihang Su\textsuperscript{\rm 1},
    Zhumin Chu\textsuperscript{\rm 1},
    Haitao Li\textsuperscript{\rm 1},
    Yujia Zhou\textsuperscript{\rm 1},
    Dingbo Yuan\textsuperscript{\rm 2},
    Xudong Wang\textsuperscript{\rm 2},
    Jun Zhou\textsuperscript{\rm 2},
    Yiqun Liu\textsuperscript{\rm 1},
    Min Zhang\textsuperscript{\rm 1},
    Shaoping Ma\textsuperscript{\rm 1},
    Qingyao Ai\textsuperscript{\rm 1}\thanks{Corresponding Author: aiqy@tsinghua.edu.cn}
}
\affiliations{
    \textsuperscript{\rm 1}Department of Computer Science and Technology, Tsinghua University
    \textsuperscript{\rm 2}Ant Group
}

\usepackage{bibentry}
\begin{document}

\maketitle

\begin{abstract}
The rapid development of large language models (LLMs) has highlighted the need for efficient and reliable methods to evaluate their performance.
Traditional evaluation methods often face challenges like high costs, limited task formats, dependence on human references, and systematic biases.
To address these limitations, we propose Auto-PRE, an automatic LLM evaluation framework inspired by the peer review process.
Unlike previous approaches that rely on human annotations, Auto-PRE automatically selects evaluator LLMs based on three core traits: consistency, pertinence, and self-confidence, which correspond to the instruction, content, and response stages, respectively, and collectively cover the entire evaluation process.
Experiments on three representative tasks, including summarization, non-factoid QA, and dialogue generation, demonstrate that Auto-PRE achieves state-of-the-art performance while significantly reducing evaluation costs.
Furthermore, the structured and scalable design of our automatic qualification exam framework provides valuable insights into automating the evaluation of LLMs-as-judges, paving the way for more advanced LLM-based evaluation frameworks. 
\end{abstract}

\begin{links}
    \link{Code}{https://github.com/cjj826/Auto-PRE}
\end{links}

\section{Introduction}
Recently, the rapid advancement of large language models (LLMs) has attracted significant attention from both academia and industry \cite{yang2025qwen3,liu2024deepseek}. As LLMs evolve rapidly, how to evaluate their performance effectively and efficiently has become a crucial question. 

Existing evaluation methods for LLMs can be categorized into two types: manual evaluation \cite{Zheng2023JudgingLW} and automatic evaluation \cite{chang2024survey}. Manual evaluation is considered the most reliable and effective method, but it is usually suboptimal due to its high costs in practice. 
Automatic evaluation aims to reduce the cost by directly assessing model performance without human annotations. 
However, existing automatic evaluation methods often support limited types of task formats (e.g., multiple-choice questions) and need human-created references for judgments. 
While recent studies have attempted to build reference-free frameworks for open-ended task evaluation with LLMs, research \cite{zeng2024eeif} has shown LLM-based evaluators (or reviewers), including the powerful GPT-4 \cite{achiam2023gpt}, may exhibit a preference for answers generated by models with the same origin. In our paper, we refer to this preference as a systematic bias, which could limit the reliability of the evaluation framework in practice.

\begin{figure}[t]
    \centering
    \includegraphics[width=\columnwidth]{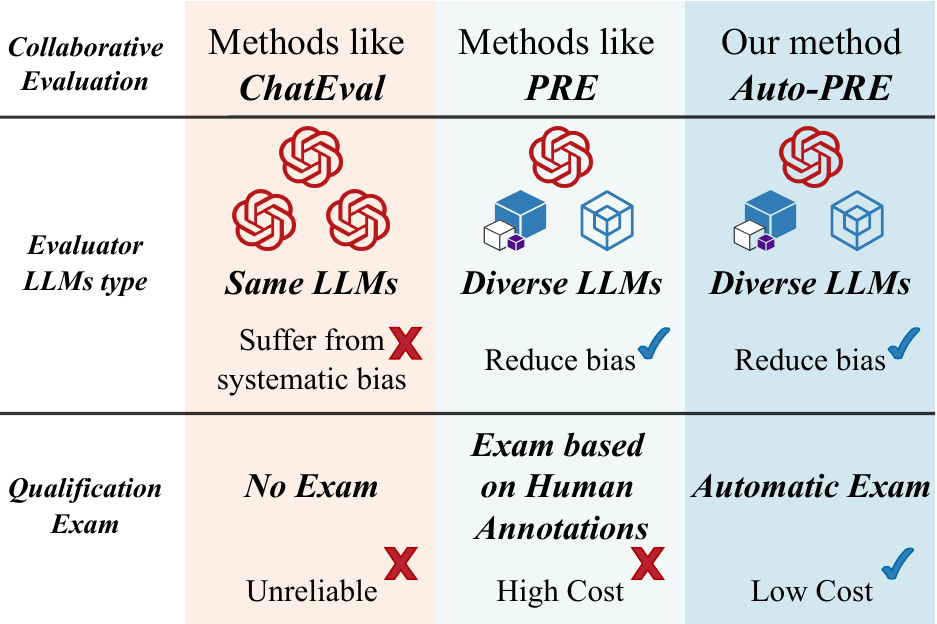}
    \caption{Comparison of existing collaborative evaluation methods. Our Auto-PRE offers advantages in reducing bias and lowering cost.}
    \label{fig:1-1}
\end{figure}

\begin{figure*}[t]
    \centering
    \includegraphics[width=0.7\textwidth]{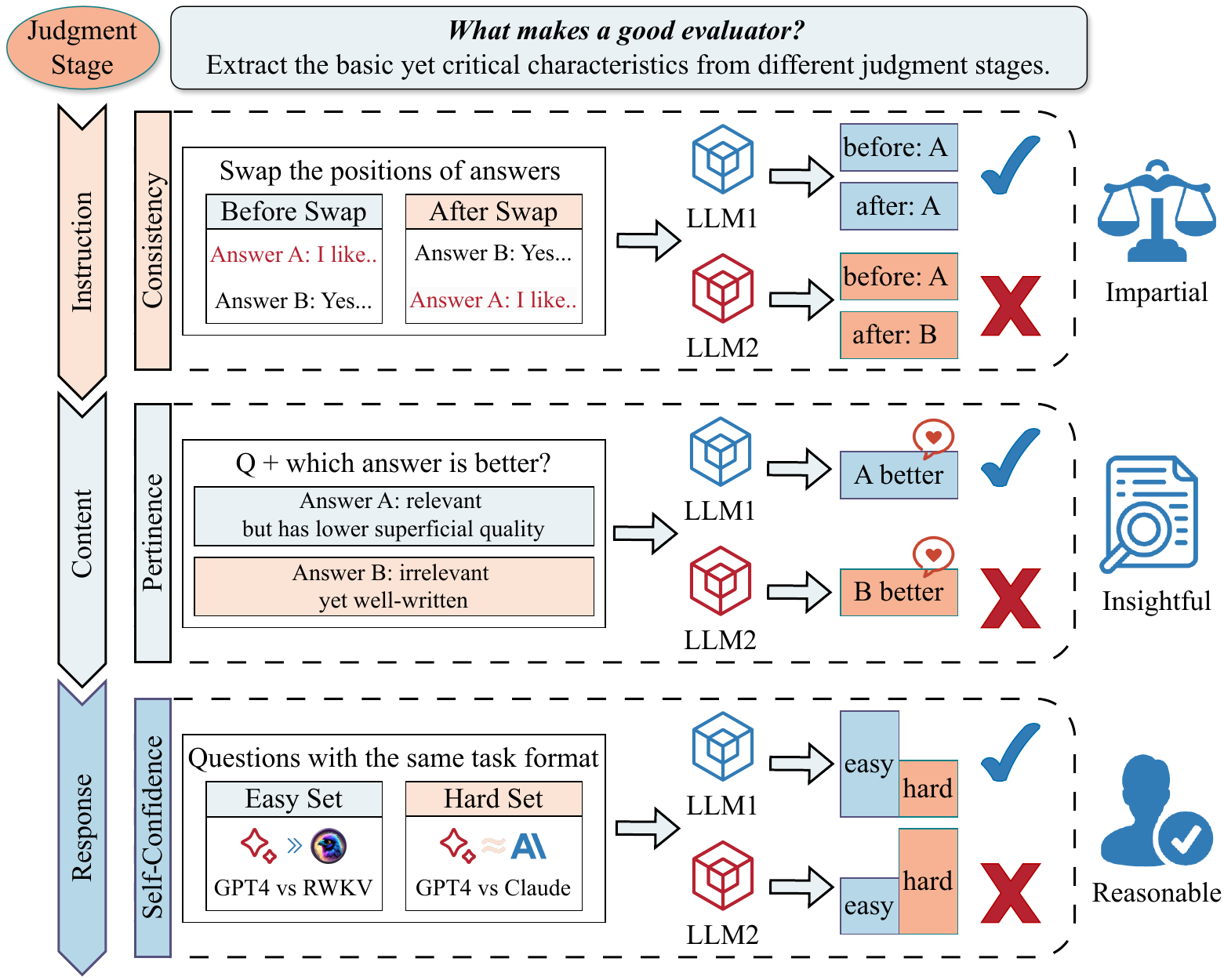}
    \caption{The framework of our automatic qualification exam. (1) \textit{Consistency} measures the proportion of consistent outputs by the LLM after swapping answer positions in prompts; (2) \textit{Pertinence} assesses whether the LLM evaluates based on the pertinence of answers to the question, unaffected by their superficial quality; (3) \textit{Self-Confidence} determines if the LLM exhibits higher confidence on easier question sets when facing two sets of the same format but objectively different difficulties.}
    \label{fig:3-1}
\end{figure*}

To develop more advanced evaluation methods, recent research has investigated the possibility of employing multiple LLMs to evaluate collaboratively, achieving notable performance. However, similar to human collaboration, when unqualified LLMs participate in collaboration, they often impair the method's performance. Therefore, selecting the appropriate LLMs as evaluators is a crucial issue. 
As shown in Figure \ref{fig:1-1}, methods like ChatEval \cite{chan2024chateval} directly select powerful LLMs such as GPT-3.5-turbo or GPT-4 to build multiple agents to debate and collaborate for evaluation. 
However, since it only utilizes LLMs from the same series, it still suffers from the systematic bias.
Another method, PRE \cite{chu2024pre}, simulates the academic peer-review mechanism by selecting qualified evaluator LLMs from a diverse pool of candidates through a qualification exam, which measures the accuracy of their outputs against human annotations. The final evaluation scores are then computed via weighted aggregation of the selected evaluators’ outputs.
Although experiments indicate PRE can reduce the  systematic bias, it still relies on human-annotated data for the exam and is therefore not fully automatic and high-cost.

In order to address the aforementioned limitations, we propose Auto-PRE, a peer-review framework that introduces an automatic qualification exam to select qualified evaluator LLMs without relying on human annotations. To achieve this, the key question we need to answer is: \textit{What makes a good evaluator?} Our guiding principle is that qualified evaluator LLMs should exhibit characteristics similar to those of excellent human evaluators. However, the inherent complexity of human evaluators makes directly summarizing their characteristics challenging. To strike a balance between rationality and feasibility, we first structure the evaluation process into three stages: Instruction (evaluation prompt), Content (material to be assessed), and Response (assessment result generated by the evaluator). This division covers the entire evaluation process and ensures the completeness of our exam. We then extract basic yet critical characteristics from each stage: (1) Consistency: Upon receiving the judgment instruction, the evaluator should have no preset biases to ensure the objectivity; (2) Pertinence: When judging the specific content, the evaluator is expected to have a thorough understanding of the task and identify the core factors that truly impact the quality of the answers (e.g., pertinence to given question), rather than relying solely on secondary or superficial factors;
(3) Self-Confidence: After providing the judgment response, the evaluator should have a reasonable confidence to reflect the reliability \cite{zhao2015imperfect}. 
Based on these characteristics, we propose three automatic methods for selecting evaluator LLMs for peer review. 
All these methods require no human annotations, making the framework fully automatic, cost-efficient, scalable, and robust to LLM-introduced systematic bias.

Experimental results on three tasks, including summary generation, non-factoid question-answering, and dialogue generation, indicate that our Auto-PRE can achieve state-of-the-art performance at a much lower cost. Furthermore, our qualification exam covers the entire evaluation process and is easily adapted to support additional traits, serving as a potential guide for automating the evaluation of LLMs-as-judges and advancing the LLM-based evaluation methods.

\section{Related Work}
Evaluation methods for LLMs fall into manual and automatic types. Manual evaluation (e.g., Chatbot Arena \cite{Zheng2023JudgingLW}) offers high reliability but is expensive and non-scalable. Automatic methods include: (1) reference-based metrics such as BLEU \cite{papineni2002bleu}, ROUGE \cite{lin2004rouge}, and BERTScore \cite{zhang2019bertscore}, which are limited in capturing answer quality, and vulnerable to overfitting if references are leaked; (2) multiple-choice evaluation, which is simple but insufficient for open-ended tasks; and (3) LLM-based evaluation~\cite{chan2024chateval,chu2024pre}, where LLMs serve as evaluators. Although LLM-based evaluation methods have shown competitive evaluation performance, these evaluator LLMs still have many flaws, such as a preference for verbose answers or answers generated by similar LLMs \cite{zeng2024eeif}. Therefore, researchers have begun to explore whether these evaluator LLMs are truly qualified. Some of this work involves constructing evaluation benchmarks through random sampling output pairs and crowdsourced manual annotations, such as LLMEval \cite{zhang2023wider}, MT-Bench \cite{Zheng2023JudgingLW}, and FairEval \cite{wang2023large}, but these works overlook biases introduced by the subjective preferences of human annotators. To address this, some researchers attempt to create more objective meta-evaluation benchmarks; for instance, LLMBAR \cite{zeng2024eeif} is designed to assess if evaluator LLMs follow instructions, with data samples that have been manually checked by experts to ensure objectivity. In contrast to these annotation-heavy evaluation methods, our methods automatically select evaluator LLMs based on objective traits across different judgment stages.

\section{Methodology}
As previously discussed, one of the key issues in collaborative evaluation methods is how to automatically, objectively, and cost-effectively select qualified evaluator LLMs. 
Based on the different judgment stages, we extract basic yet critical characteristics: Consistency, Pertinence, and Self-Confidence. 
Leveraging these three characteristics, we have designed an automatic qualification exam that incorporates three different selection methods, as shown in Figure \ref{fig:3-1}. 
It is worth noting that the ability requirements for evaluators may vary across different tasks or even between datasets within the same task. However, the requirements tend to be similar for different instances within the same dataset of a given task. To balance effectiveness and cost-efficiency, our selection methods are designed to be performed per dataset. Next, we provide a detailed description of our methods.

\subsection{Consistency}
When receiving judgment instructions, an excellent evaluator should remain objective, impartial, and consistent, avoiding the influence of preset biases. Consider a task to evaluate the quality of $n$ answers, denoted as $\{Y_1, Y_2, \dots, Y_n\}$, to the same question $Q$. Within the judgment instructions, there exists a subset that should not affect the evaluation outcomes, denoted as $NI$. This subset includes factors such as the order of answers and the placement of question-answer pairs within the prompt. A qualified evaluator should demonstrate no bias towards specific $NI$. Formally, this means the evaluation score for any given answer $Y_t$ should remain invariant under changes to $NI$:
\begin{equation}\label{eqn-3.1} 
Score(Y_t \mid NI_1) = Score(Y_t \mid NI_2),
\end{equation}
where $NI_1$ and $NI_2$ represent different $NI$.

Currently, numerous studies \cite{wang2023large, li2024calibraeval} have shown that some LLMs exhibit various preset biases, so we can use the degree of biases of different LLMs for selection. Specifically, we implement the most common position bias and set $n = 2$. The candidate LLM \( L \) is given the tuple \( (Q, Y_1, Y_2) \) as input, and generates a preference relation \( T_1 \). Next, the positions of \( Y_1 \) and \( Y_2 \) are swapped to form the tuple \( (Q, Y_2, Y_1) \), which is then inputted to \( L \) to generate another preference relation \( T_2 \). We randomly sample \( m \) instances from the specific dataset, with \( (Q_i, Y_{1,i}, Y_{2,i}) \) (\( i = 1, 2, \dots, m \)). Then, the proportion of consistent outputs for the candidate LLM \( L \) is computed as:
\begin{equation}\label{eqn-31} 
P_c = \frac{\sum_{i=1}^{m} \mathbb{I}(T_{1,i} = T_{2,i})}{m},
\end{equation}
where \( \mathbb{I}(\cdot) \) is the 0-1 indicator function, and this definition remains consistent throughout our paper. If the \( P_c \) exceeds the threshold \( \eta_c \), the candidate LLM \( L \) is considered to have passed the exam. It should be noted that our framework is highly scalable and can be easily adapted to support the detection of other types of biases based on task requirements, such as token or prompt bias.

\begin{table}[t]
\centering
\begin{tabular}{p{0.4\columnwidth}p{0.5\columnwidth}}
\toprule
\textbf{Q} & \textbf{Q'} \\
\midrule
  Could someone define \textbf{Christian} for me? & 
Can anyone explain the concept of \textbf{Buddhism} to me?
\\
\midrule
What is the best Fantasy \textbf{Football} Platform?
 &
What is the top Fantasy \textbf{Baseball} App?
 \\
\bottomrule
\end{tabular}
\caption{Some cases of how GPT-4 modifies Q to Q'.}
\label{tab:3-2}
\end{table}

\subsection{Pertinence}
\label{section_3.3}
When judging specific contents, unqualified evaluators may struggle to identify the core factors that truly impact answer quality (e.g., the pertinence between answer and question)~\cite{zeng2024eeif}. Instead, they tend to provide unreliable scores based on secondary or superficial factors such as the length or format of the answer. Therefore, we design a method to select insightful evaluator LLMs based on whether the candidate LLMs can effectively distinguish between the answer's pertinence to the given question and its superficial quality. To implement this method, we generate two types of answers: (1) answers that are highly pertinent to the given questions but of lower superficial quality, denoted as RA (Relevant Answers); and (2) answers that are less pertinent but exhibit higher superficial quality, denoted as IA (Irrelevant yet well-written Answers). Specifically, the process of constructing RA and IA involves two steps:

\textbf{Step 1:} Generate a variant of the original question $Q$, denoted as $Q'$, where $Q$ and $Q'$ are similar but sufficiently different to ensure that answers generated based on each have significantly different pertinence to the original question $Q$. The difference in pertinence decreases as the similarity between $Q$ and $Q'$ increases. We design two methods for constructing $Q'$: one is to search other questions from the same dataset as $Q$ to find a suitable $Q'$, and the other is to prompt a capable LLM (like GPT-4) to modify $Q$ to obtain $Q'$. Table \ref{tab:3-2} shows that GPT-4 mainly achieves the transformation from $Q$ to $Q'$ by changing the keywords in the $Q$.

\textbf{Step 2:} Select one LLM's response to $Q$ as the RA and another LLM's response to $Q'$ as the IA. Here, the LLM generating the IA can be a more capable LLM than the LLM generating the RA, or it can be the candidate LLM itself. The former is based on the assumption that a more capable LLM is likely to produce answers with higher superficial quality, while the latter assumes that the candidate LLM considers its own answers to be of sufficient superficial quality. 

After obtaining $m$ pairs of RA and IA, we calculate the proportion of candidate LLM outputs where the $RA_i$ is rated better than the $IA_i$ as a filtering metric:
\begin{equation}\label{eqn-3.2} 
P_p = \frac{\sum_{i=1}^{m} \mathbb{I}(RA_i > IA_i)}{m}.
\end{equation}
If $P_p$ exceeds a certain threshold $\eta_p$, the candidate LLM is considered qualified.
\subsection{Self-Confidence}
\label{section 3.2.3}
After providing judgment responses, like reliable human evaluators, qualified evaluator LLMs should have a reasonable self-confidence level based on their understanding of the task difficulty and their capabilities~\cite{zhao2015imperfect}. As for what counts as a reasonable level of self-confidence, a suitable prior assumption is that when the same LLM encounters two questions with the same task format but objectively different difficulties, it should have more self-confidence in solving the easier question. It is noteworthy that in the above assumption, the task formats for both questions must be identical to ensure that the LLM requires the same capabilities to solve them. Additionally, the difference in difficulties between the two questions must be based on objective criteria rather than human subjective judgment, to eliminate potential biases arising from the disagreement between humans and LLMs.

Based on this assumption, we select those LLMs that show higher self-confidence on the easier set than on the harder ones as evaluators. Implementing this requires addressing two key issues:
(1) How to construct two question sets with the same task format but objectively different difficulties?
(2) How to extract the self-confidence of LLMs?

\begin{table}[t]
\begin{tabular}{p{0.2\columnwidth}p{0.7\columnwidth}}
\toprule
 & \textbf{Content} \\
\midrule
Original News & The man's body was discovered in a field near Belsyde Avenue... \\
\midrule
GPT-4's Summary & A man's body was found in a field near Belsyde Avenue... \\
\midrule
Claude's Summary & The body of an unidentified man was discovered in a field near... \\
\midrule
RWKV's Summary & Bob: Hey Alice, have you heard about the death of a man... \\
\bottomrule
\end{tabular}
 \caption{Summaries by different LLMs.}
\label{tab:3-1}
\end{table}

Regarding issue (1), we initially select evaluating the quality differences between two answers to the same question as our task format. For this task, an objective principle holds: the more similar the two answers in quality, the more difficult it becomes to distinguish between them. Based on this, we construct the easy and hard question sets by assuming that the similarity in answer quality is correlated with the capability gap between the two LLMs that generate them. Specifically, we pair LLMs with a large capability gap (e.g., GPT-4 vs. RWKV-7B) to form the easy set, and LLMs with similar capabilities (e.g., GPT-4 vs. Claude) to form the hard set. Table \ref{tab:3-1} presents a summarization example where GPT-4 and Claude generate valid summaries, while RWKV-7B produces an irrelevant dialogue. This highlights that comparing high-quality answers (e.g., GPT-4 vs. Claude) is objectively more challenging than answers with a clear quality gap (e.g., GPT-4 vs. RWKV-7B).

Regarding issue (2), we convert the probability of outputting a specific token into the uncertainty of the output \cite{duan2023shifting,manakul2023selfcheckgpt}, and assume that higher uncertainty represents lower self-confidence. Specifically, for the task we select, which involves evaluating the quality difference between two answers, the candidate LLMs are only required to output the specific token `one' or `two' to indicate which answer is better. Through this simplification, we can directly convert the probability $p$ of an LLM outputting `one' or `two' into the LLM’s uncertainty $-log(p)$, thereby obtaining self-confidence. However, this method requires access to the probabilities of the specific tokens, making it unsuitable for some closed-source LLMs. For this situation, we directly prompt these LLMs to output specific self-confidence level labels. Since these popular and closed-source LLMs (e.g., GPT-4) typically have an enormous number of parameters and exceptional capabilities, this straightforward and low-cost method shows good performance in our experiments.

If the candidate LLM's average confidence level on the easy set is \( S_{\text{easy}} \), and its average confidence level on the hard set is \( S_{\text{hard}} \), then $P_{s}$ is defined as follows:

\begin{equation}\label{eqn-33} 
P_s =
\begin{cases} 
1, & \text{if } S_{\text{easy}} > S_{\text{hard}} \\
0, & \text{otherwise}.
\end{cases}
\end{equation}
More details on the implementation and validation of our methods' effectiveness are in Appendix Section 3.

\begin{table*}[t]
\centering
\setlength{\tabcolsep}{1mm}  
\begin{tabular}{c|ccc|ccc|ccc}
\hline
Methods                                                               & \multicolumn{3}{c|}{Xsum}                                                                           & \multicolumn{3}{c|}{NF\_CATS}                                                                                & \multicolumn{3}{c}{DailyDialog}                                                                     \\ \hline
ROUGE-L                                                               & \multicolumn{3}{c|}{0.5798}                                                                         & \multicolumn{3}{c|}{/}                                                                                       & \multicolumn{3}{c}{0.6984}                                                                               \\
\begin{tabular}[c]{@{}c@{}}BERTScore\end{tabular}        & \multicolumn{3}{c|}{0.5901}                                                                         & \multicolumn{3}{c|}{/}                                                                                       & \multicolumn{3}{c}{0.6143}                                                                               \\
\begin{tabular}[c]{@{}c@{}}GPTScore\end{tabular} & \multicolumn{3}{c|}{0.6910}                                                                         & \multicolumn{3}{c|}{0.5940}                                                                                  & \multicolumn{3}{c}{0.3445}  \\ \hline
                         & 5-level                            & 100-level                 & pairwise                           & 5-level                            & 100-level                          & pairwise                           & 5-level                   & 100-level                          & pairwise                           \\ \hline
Vicuna-7b-v1*             & 0.5100                             & 0.5215                    & 0.5106                             & 0.5503                             & 0.5735                             & 0.5036                             & 0.5712                    & 0.5302                             & 0.5000                             \\
ChatGLM3-6B*              & 0.6564                             & 0.6288                    & 0.5127                             & 0.5518                             & 0.5551                             & 0.5244                             & 0.5806                    & 0.6093                             & 0.5213                             \\
Baichuan2-13b*            & 0.5745                             & 0.6247                    & 0.6057                             & 0.5521                             & 0.5500                             & 0.5515                             & 0.5483                    & 0.6055                             & 0.6260                             \\
FastChat-t5-3b*           & 0.6180                             & 0.6553                    & 0.6921                             & 0.5411                             & 0.5708                             & 0.6537                             & 0.5669                    & 0.5759                             & 0.6614                             \\
GPT-3.5-turbo*            & 0.6840                             & 0.6695                    & 0.6470                             & 0.5586                             & 0.5592                             & 0.6080                             & 0.6814                    & 0.6542                             & 0.6812                             \\
ChatGLM-Pro*              & 0.6553                             & 0.7033                    & 0.6951                             & 0.6485                             & 0.6887                             & 0.7042                             & 0.6001                    & 0.6497                             & 0.7412                             \\
GPT-4*                    & 0.6893                             & 0.7005                    & 0.7369                             & 0.6330                             & 0.6801                             & \( 0.7815^{\dagger\dagger} \)      & 0.6732                    & 0.6752                             & 0.8088                             \\ \hline
PandaLM                                                              & /                                  & /                         & 0.6350                             & /                                  & /                                  & 0.7205                             &   /                        &   /                                 &    0.7039                                \\
DeepSeek-R1-0528                                                              & 0.6809                                  & 0.7131                         & 0.7119                             & 0.6589                                 & 0.7095                                  & 0.7159                             &   0.5923                        &   0.6668                               &    0.7742                                \\
ChatEval                 & 0.5694                             & 0.5747                    & 0.6584                             & 0.6009                             & 0.6435                             & 0.7366                             & 0.6080                    & 0.6725                             & 0.6820                             \\
PRE (w/o Filter)         & 0.7055                             & 0.7002                    & 0.7401                             & 0.6804                             & 0.6711                             & 0.7542                             & 0.7258                    & 0.7295                             & 0.7413                             \\
PRE (Auto-Exam)          & 0.7064                             & 0.7133                    & 0.7381                             & 0.6795                             & 0.6905                             & 0.7664                             & 0.7248                    & 0.7129                             & 0.8048                             \\ \hline
PRE (Human Filter)       & $0.7211^{\dagger\dagger}$          & 0.7192                    & 0.7423                             & 0.6824                             & $0.7104^{\dagger\dagger}$          & $0.7801^{\dagger\dagger}$          & 0.7255                    & $0.7318^{\dagger\dagger}$          & 0.8085                             \\ \hline
Auto-PRE (ours)                 & $\textbf{0.7231}^{\dagger\dagger}$ & $\textbf{0.7195}^\dagger$ & $\textbf{0.7462}^{\dagger\dagger}$ & $\textbf{0.6887}^{\dagger\dagger}$ & $\textbf{0.7146}^{\dagger\dagger}$ & $\textbf{0.7821}^{\dagger\dagger}$ & $\textbf{0.7305}^\dagger$ & $\textbf{0.7469}^{\dagger\dagger}$ & $\textbf{0.8161}^{\dagger\dagger}$ \\ \hline
\end{tabular}
\caption{The overall performance (accuracy) of Auto-PRE and other baselines. The best result is highlighted in bold. $\dagger$/$\dagger\dagger$ indicates $p$-value of paired sample t-test where the method outperforms PRE (Auto-Exam) is less than 0.05/0.01.}
\label{tab:5-5}
\end{table*}
\section{Experimental Setup}
\subsection{Tasks And Datasets}
\label{section 4.1}
Unlike the automatic evaluation methods for multiple-choice questions, we focus on open-ended questions. To this end, we select three generative tasks along with representative datasets: (1) Summary generation: Xsum \cite{narayan-etal-2018-dont}, (2) Non-factual question-answering: NF\_CATS \cite{bolotova2022non}, (3) Dialogue generation:  DailyDialog \cite{li2017dailydialog}. For the above three datasets, we randomly sample 100 instances from each as the question set and select 7 different LLMs to generate answers, forming the answer set. Then, we employ human annotators to provide annotations over the answer set. We use these annotations as ground truth to evaluate the performance of each evaluation method. More dataset details can be found in Appendix Section 1.

\subsection{Evaluation Formats And Metrics}

We compare two evaluation formats: pointwise and pairwise. For pointwise, we design two implementation approaches: 5-level and 100-level. For pairwise evaluation, we minimize the bias introduced by the positioning of answers by calculating the mean of the evaluation results before and after swapping the positions of the two answers. We use accuracy as the main evaluation metric, defined as the agreement rate between the manual preference annotations and the method's evaluation results. When using pointwise evaluation, we will convert each answer's individual scores into pairwise rankings. In addition, for pointwise, we also use Spearman correlation coefficient \cite{lehman2013jmp} ($S$) to measure the consistency between the model’s outputs and manual annotations. We calculate $S$ for each instance of the task, and report the mean of them as the overall performance.

\subsection{Baselines}
We compare our Auto-PRE with different baselines:

\noindent\textbf{1. ROUGE-L and BERTScore}: These are widely used reference-based metrics (not applicable to the NF\_CATS due to the lack of reference answers). For BERTScore, we use deberta-xlarge-mnli~\cite{he2020deberta} as the base model. We report the F1 score for both metrics.

\noindent\textbf{2. GPTScore}~\cite{fu2023gptscore}: This metric assesses the quality of generated text by computing the log-probability of the output given the prompt under a specific LLM. We use text-davinci-003~\cite{brown2020language} as the base model.

\noindent\textbf{3. Single LLM}: Use a single general LLM as evaluator. The LLMs used are marked with an asterisk (*) in Table \ref{tab:5-5}.

\noindent\textbf{4. PandaLM}~\cite{wang2024pandalm}: A fine-tuned variant of LLaMA-7B specifically designed for preference judgment.

\noindent\textbf{5. Deepseek-R1-250528}~\cite{guo2025deepseek}: A strong representative of recently popular reasoning-specialized LLMs.

\noindent\textbf{6. ChatEval}~\cite{chan2024chateval}: Use two GPT-3.5-turbo to build two agents as evaluators to debate in two rounds with a one-by-one communication strategy.

\noindent\textbf{7. PRE (Human Filter)}: Use all 7 LLMs in Single LLM as candidates and select evaluators based on human annotations. Unless noted otherwise, PRE refers to this version.

\noindent\textbf{8. PRE (w/o Filter)}: Use all candidates as evaluators without any filtering.

\noindent\textbf{9. PRE (Auto-Exam)}: The original PRE offers a preliminary and simplified version of the automatic qualification exam, using only consistency for selection.

\subsection{Implementation Details}
We set \textit{temperature} to 0 and \textit{do\_sample} to False for reproducibility. In our Auto-PRE, \( \eta_c \) and \( \eta_p \) are respectively defined as the mean $P_c$ and $P_p$ across all candidate LLMs. The fusion weight for each evaluator LLM is the average of $P_c$, $P_p$, and $P_s$. Evaluation prompts and hyperparameter details are in Appendix Sections 2 and 5, respectively.

\section{Experimental Results}
In this section, we present the experimental results and aim to address the following four research questions (RQs):

\noindent 1. How does the performance of Auto-PRE compare to other baseline methods?

\noindent 2. Does Auto-PRE help mitigate the systematic bias caused by relying on LLMs from the same series?

\noindent 3. What are the benefits of Auto-PRE in reducing costs?

\noindent 4. How do the three selection methods in Auto-PRE interact and complement each other?
\subsection{Main Results (RQ1)}
\begin{table}[t]
\small
\setlength{\tabcolsep}{0.8mm}
\begin{tabular}{ccccccc}
\hline
Methods                                                       & \multicolumn{2}{c}{Xsum}          & \multicolumn{2}{c}{NF\_CATS}      & \multicolumn{2}{c}{DailyDialog}   \\ \hline
ROUGE-L                                                       & \multicolumn{2}{c}{0.2329}        & \multicolumn{2}{c}{/}             & \multicolumn{2}{c}{0.4057}             \\
BertScore                                                     & \multicolumn{2}{c}{0.2715}        & \multicolumn{2}{c}{/}             & \multicolumn{2}{c}{0.5137}             \\
GPTScore                                                      & \multicolumn{2}{c}{0.4203}        & \multicolumn{2}{c}{0.1966}        & \multicolumn{2}{c}{0.4589}        \\ \hline
                                                              & 5-l             & 100-l           & 5-l             & 100-l           & 5-l             & 100-l           \\ \hline
GPT-4                   & 0.4801          & 0.4701          & 0.3318          & 0.3287          & 0.5044          & 0.4684          \\
ChatEval                 & 0.1767          & 0.1942          & 0.2142          & 0.2560          & 0.3865          & 0.4333          \\ \midrule
\begin{tabular}[c]{@{}c@{}}PRE \\ (w/o Filter)\end{tabular}   & 0.4650          & 0.4312          & 0.3555          & 0.3103          & 0.5297          & 0.5294          \\
\begin{tabular}[c]{@{}c@{}}PRE \\ (Auto-Exam)\end{tabular}    & 0.4809          & 0.4595          & 0.3729          & 0.3515          & 0.5342          & 0.5229          \\
\begin{tabular}[c]{@{}c@{}}PRE \\ (Human Filter)\end{tabular} & 0.4991          & 0.4633          & \textbf{0.3939} & 0.3706          & 0.5347          & 0.5547          \\ \midrule
Auto-PRE                 & \textbf{0.5087} & \textbf{0.4937} & 0.3931          & \textbf{0.3818} & \textbf{0.5382} & \textbf{0.5599} \\ \hline
\end{tabular}%

\caption{The spearman correlation coefficient ($S$) of Auto-PRE and other baselines. Here, 5-l and 100-l denote 5-level and 100-level annotations, respectively.}
\label{table: s}
\end{table}

Table \ref{tab:5-5} and \ref{table: s} show the overall results, leading to the following observations:

Across various settings, Auto-PRE consistently outperforms existing methods and achieves the best overall performance on average. In contrast, reference-based metrics such as ROUGE-L and BERTScore exhibit significant performance gaps relative to the top-performing methods, while GPTScore also falls notably behind. PandaLM performs reasonably well, but only on the NF\_CATS. Deepseek-R1-0528 demonstrates strong competitiveness, particularly under the pointwise format, highlighting the potential of reasoning LLMs for evaluation tasks. Additionally, the comparison with ChatEval in Table \ref{tab:5-5} is not entirely fair (as we don't use GPT-4 as the base for the agent due to cost considerations), but we will provide a detailed comparison of the performance between Auto-PRE and ChatEval under equivalent cost conditions in the cost analysis section.

Compared to GPT-4, Auto-PRE exhibits significant improvements in pointwise format, with an average increase of 4.53\% in accuracy. In terms of pairwise, Auto-PRE performs comparably with GPT-4. Moreover, we believe that GPT-4's systemic bias may not be apparent in the context of the overall dataset, as 70\% of the answer pairs in our experimental setup do not include LLMs from the GPT series. In the bias analysis section, we will test on instances containing LLMs from the GPT series to further discuss the advantages of Auto-PRE in mitigating bias compared to GPT-4.

Compared to PRE (w/o Filter) and PRE (Auto-Exam), Auto-PRE achieves significantly better performance while keeping low costs, with an average improvement of 2.44\% in accuracy and 0.0325 in $S$ over PRE (w/o Filter), and an average improvement of 1.45\% in accuracy and 0.0256 in $S$ over PRE (Auto-Exam). This underscores the necessity of our more well-designed qualification exam.

Compared to PRE (Human Filter), Auto-PRE achieves comparable performance while significantly reducing costs, which will be discussed in the cost analysis section. Interestingly, Auto-PRE even outperforms PRE (Human Filter) to some extent, which might be because of the more comprehensive coverage of different judgment stages in our automatic qualification exam. In contrast, while manual annotation-based filtering emphasizes evaluator accuracy, it does not fully account for all judgment stages. This will be discussed in detail in the ablation study section.

\subsection{Bias Analysis (RQ2)}
To demonstrate that Auto-PRE effectively reduces bias in single-evaluator LLM (e.g., GPT-4), we conduct experiments using GPT-3.5-turbo and ChatGLM2-6B as two answer generators, tested on the Xsum and NF\_CATS datasets in the pairwise evaluation format. 

\begin{table}[t]
\setlength{\tabcolsep}{0.8mm}
\begin{tabular}{ccccc}
\hline
\multirow{2}{*}{Methods}            & \multicolumn{2}{c}{Xsum}          & \multicolumn{2}{c}{NF\_CATS}      \\
                                    & accuracy         & rate (\%)      & accuracy         & rate (\%)      \\ \hline
GPT-4                               & 0.5366           & \textbf{83.64} & 0.8618           & \textbf{87.88} \\
PRE (w/o Filter)                    & 0.5610           & 68.18          & 0.8553           & 84.85          \\
\multicolumn{1}{l}{PRE (Auto-Exam)} & 0.5671           & 65.45          & 0.8750           & 86.36          \\
PRE (Human Filter)                  & 0.5549           & 77.27          & $\textbf{0.8816}^{*}$ & 71.21          \\
Auto-PRE                            & $\textbf{0.5854}^{*}$ & 65.45          & $\textbf{0.8816}^{*}$ & 74.24          \\ \hline
\end{tabular}%

\caption{The bias analysis of GPT-4 (pairwise). * indicates p-value of paired sample t-test, where the method outperforms GPT-4 is less than 0.05.}

\label{tab:5-6}
\end{table}

Table \ref{tab:5-6} presents the results. As Equation \ref{eqn-1} shows, $rate$ is defined as the proportion of instances where, despite human annotators judging the two LLMs as tied or preferring ChatGLM2-6B, the method still favors GPT-3.5-turbo.

\begin{equation}\label{eqn-1} 
rate_m = \frac{\sum \mathbb{I}(T_h \in \{0, -1\} \land T_m = 1)}{\sum \mathbb{I}(T_h \in \{0, -1\})} \times 100\%,
\end{equation}
where \( T_h \) and \( T_m \) denote human and method preferences, respectively; \( 1 \) indicates a preference for GPT-3.5-turbo, \( -1 \) for ChatGLM2-6B, and \( 0 \) for a tie. 

From the results, we can observe a significant performance gap between GPT-4 and Auto-PRE, with an average difference of 3.43\% in accuracy. Furthermore, GPT-4 demonstrates a notably higher $rate$ compared to Auto-PRE,  with an average disparity of 15.92\% in $rate$. This suggests that GPT-4's preference for GPT-3.5-turbo could compromise its performance and reliability. In contrast, Auto-PRE improves both overall performance and reliability by effectively leveraging collaboration across diverse LLMs.

\subsection{Cost Analysis (RQ3)}
In this section, we analyze the cost-effectiveness of Auto-PRE and compare its performance with ChatEval under equivalent costs. To achieve this, we implement several variants of the above methods. ChatEval includes three variants (C1, C2, C3), while Auto-PRE includes five variants (A1, A2, A3, A4, A5). Their detailed configurations are provided in Appendix Section 4. We use pairwise as the evaluation format on Xsum and NF\_CATS. Each task has 4200 instances, and each instance has about 1K tokens, so completing each task requires approximately 4.2 M tokens. Based on the official pricing released \cite{glmPricing, gpt3Pricing}, the costs of ChatGLM\_Pro and GPT-3.5-turbo are estimated to be similar at \$1 per million tokens. The cost of GPT-4 is estimated at \$40 per million tokens. Open-source LLMs are considered cost-free. Additionally, the cost of the qualification exam of PRE based on human annotations is about \$115 while the cost of our automatic qualification exam (less than \$1) can be neglected compared to the total costs.

% \begin{figure}[!htbp]
%   \centering
%   \includegraphics[width=\columnwidth,height=.38\textheight,keepaspectratio,pagebox=cropbox]{pics/Xsum-13.pdf}

%   \vspace{0.35\baselineskip}

%   \includegraphics[width=\columnwidth,height=.38\textheight,keepaspectratio,pagebox=cropbox]{pics/NFQA-13.pdf}
%   \caption{Pairwise performance: Xsum (top) and NF\_CATS (bottom).}
%   \label{fig:cost-two-in-one}
% \end{figure}
\begin{figure}[t]
    \centering
    \includegraphics[width=\columnwidth]{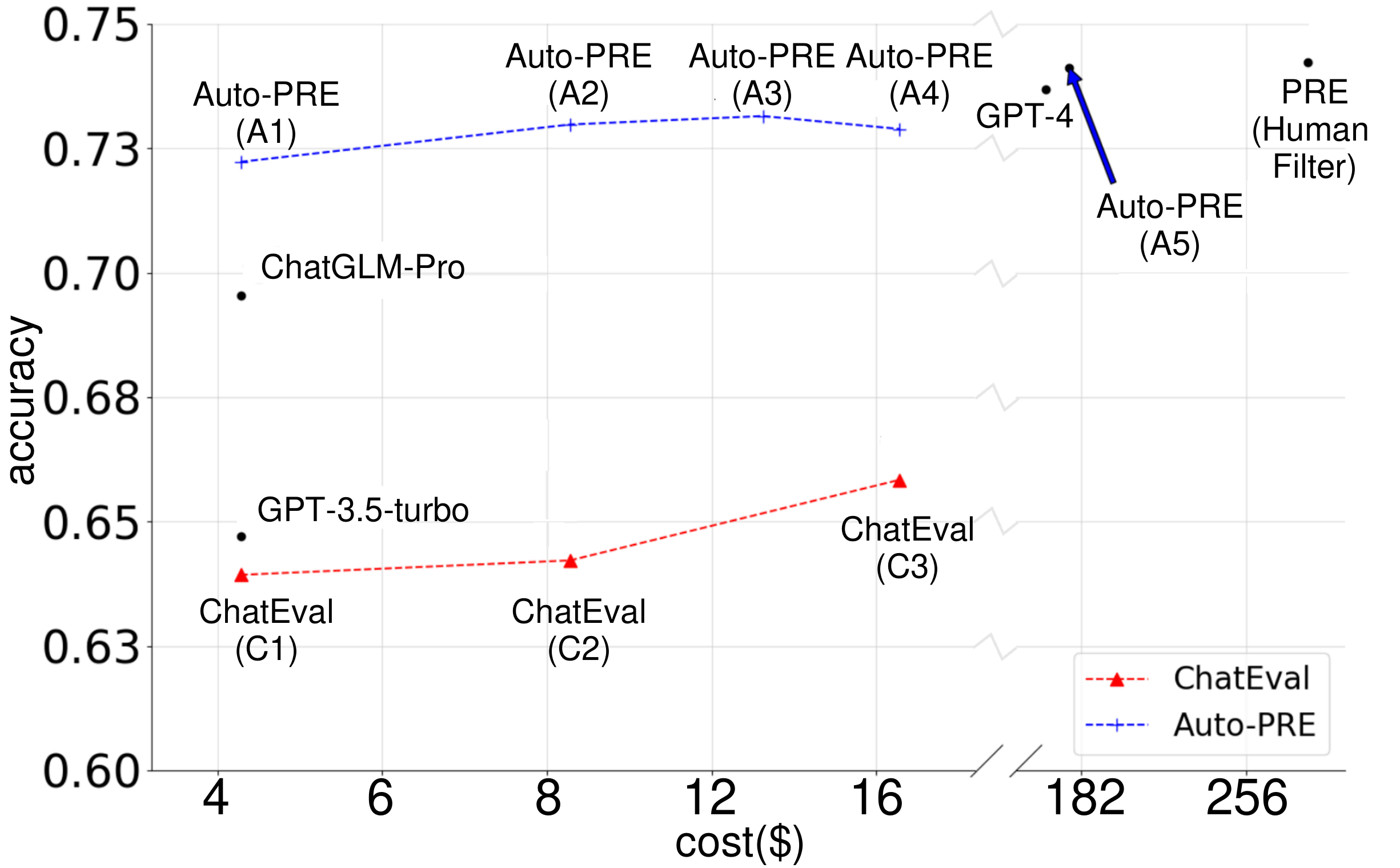}
    \caption{The performance on the Xsum (pairwise).}
    \label{fig:5-2}
\end{figure}

Figure \ref{fig:5-2} shows the relationship between the total cost and accuracy. The results show that at the same cost, our Auto-PRE (A1, A2, A3, A4, A5) can achieve higher performance than all baselines, including ChatEval (C1, C2, C3). Compared to PRE (Human Filter), our methods significantly reduce costs (about \$115) while maintaining performance without notable differences. Compared to GPT-4, our methods can reduce costs by 90\%, while keeping nearly the same performance (with only a 0.54\% decrease in accuracy). Overall, our methods can achieve performance comparable to state-of-the-art methods at a much lower cost.

\subsection{Ablation Studies (RQ4)}
In this section, we delve into the contributions of each selection method of Auto-PRE to the overall performance. We explore the performance of several variants of Auto-PRE that combine different selection methods on three tasks in the pairwise task format, including PRE (Auto-Exam): using only Consistency; Auto-PRE (P): using only Pertinence; Auto-PRE (S): using only Self-Confidence; and Auto-PRE (C+P+S): using all three methods.

\begin{table}[t]
% \small
\setlength{\tabcolsep}{1mm} 
\begin{tabular}{l|l|l|l}
\hline
\multirow{2}{*}{Methods}                                      & \multicolumn{1}{c|}{Xsum}                                                                          & \multicolumn{1}{c|}{NF\_CATS}                                                                  & \multicolumn{1}{c}{DailyDialog}                                                                  \\ \cline{2-4} 
                                                              & acc [pass]                          & acc [pass]                                                                        & acc [pass]                                                                           \\ \hline

PRE (Auto-Exam)   & \begin{tabular}[c]{@{}l@{}}0.7381\\ {[}4,5,6,7{]}\end{tabular}                             & \begin{tabular}[c]{@{}l@{}}0.7664\\ {[}5,6,7{]}\end{tabular}                           & \begin{tabular}[c]{@{}l@{}}0.8048\\ {[}4,5,6,7{]}\end{tabular}                           \\ \midrule
Auto-PRE (P)                                                  & \begin{tabular}[c]{@{}l@{}}0.7379\\ {[}2,4,5,6,7{]}\end{tabular}                           & \begin{tabular}[c]{@{}l@{}}0.7702\\ {[}4,5,6,7{]}\end{tabular}                         & \begin{tabular}[c]{@{}l@{}}0.8065\\ {[}3,4,5,6,7{]}\end{tabular}                         \\ \midrule
Auto-PRE (S)                                                  & \begin{tabular}[c]{@{}l@{}}0.7398\\ {[}3,4,5,6,7{]}\end{tabular}                            & \begin{tabular}[c]{@{}l@{}}0.7658\\ {[}3,4,6,7{]}\end{tabular}                         & \begin{tabular}[c]{@{}l@{}}0.7900\\ {[}2,5,6,7{]}\end{tabular}                           \\ \midrule
PRE (Human Filter) & \begin{tabular}[c]{@{}l@{}}0.7423\\ {[}3,4,5,6,7{]}\end{tabular}                           & \begin{tabular}[c]{@{}l@{}}$0.7801^{\dagger\dagger}$\\ {[}4,5,6,7{]}\end{tabular}      & \begin{tabular}[c]{@{}l@{}}0.8085\\ {[}5,6,7{]}\end{tabular}                             \\ \midrule
Auto-PRE (C+P+S)   & \begin{tabular}[c]{@{}l@{}}$\textbf{0.7462}^{\dagger\dagger}$\\ {[}4,5,6,7{]}\end{tabular} & \begin{tabular}[c]{@{}l@{}}$\textbf{0.7821}^{\dagger\dagger}$\\ {[}6,7{]}\end{tabular} & \begin{tabular}[c]{@{}l@{}}$\textbf{0.8161}^{\dagger\dagger}$\\ {[}5,6,7{]}\end{tabular} \\ \hline
\end{tabular}
\caption{The performance (accuracy) of different Auto-PRE variants (pairwise). The meaning of $\dagger\dagger$ is the same as in Table \ref{tab:5-5}. `pass' records the qualified LLMs in specific exams (Vicuna-7b-v1, ChatGLM3-6B, Baichuan-2-13b, FastChat-t5-3b, GPT-3.5-turbo, ChatGLM-Pro, and GPT-4 are abbreviated as integers 1-7).}
\label{tab18}
\end{table}

The results in Table \ref{tab18} indicate that Auto-PRE (C+P+S), which integrates all three selection methods, achieves the best performance, significantly outperforming PRE (Auto-Exam), Auto-PRE (P), and Auto-PRE (S) that use only a single selection method, with an average improvement of 1.33\% in accuracy. This demonstrates that the three selection methods exhibit a synergistic effect, complementing each other to enhance overall performance.

Interestingly, we observe Auto-PRE (C+P+S) even outperforms PRE (Human Filter) to some extent, showing the potential of our automatic methods. Taking NF\_CATS as an example, in our automatic qualification exam, GPT-3.5-turbo is filtered out in the self-confidence test due to exhibiting unreasonable confidence levels. In contrast, this issue is not detected by the manual annotation-based qualification exam. This suggests our method covers a broader range of judgment stages: it not only considers the model's unbiased understanding of the judgment instruction but also emphasizes its insightful comprehension of the judgment content and its reasonable self-confidence level after generating a judgment response. In comparison, the manual annotation-based qualification exam places greater emphasis on the accuracy of evaluation results while neglecting aspects such as the judge's self-confidence in its generated responses.

\section{Conclusion}
This paper develops the Auto-PRE by designing an automatic qualification exam based on three characteristics: (1) Consistency, (2) Pertinence, (3) Self-Confidence extracted from different judgment stages. Experimental results indicate that Auto-PRE achieves state-of-the-art performance while significantly reducing cost. By providing a scalable and efficient qualification exam, our work lays a foundation for automating the evaluation of LLMs-as-judges and improving the reliability of LLM-based evaluation methods.

\bibliography{aaai2026}

\section{Appendix}
\subsection{1. Dataset Details}

Unlike the automatic evaluation methods for multiple-choice questions, we focus on open-ended questions. To this end, we select three representative generative tasks:

\textbf{Summary generation} involves generating a summary for a given text. For this task, we utilize the Extreme Summary (Xsum) dataset \cite{narayan-etal-2018-dont}, which consists of over 220,000 real single-document news and summaries.

\textbf{Non-factual question-answering} refers to providing answers to questions that do not have fixed responses. For this task, we choose the NF\_CATS dataset \cite{bolotova2022non}, which contains about 12,000 non-factual questions.

\textbf{Dialogue generation} requires generating human-like responses to numerous topics in daily conversation contexts. For this, we select DailyDialog \cite{li2017dailydialog}, which is a high-quality dataset of 13k multi-turn dialogues.

For Xsum and NF\_CATS, we reuse the manual preference annotations over 7 LLMs' outputs from the PRE, which can be found in Table \ref{tab:4-1}.  We also use the experimental settings from the original PRE, utilizing the answers and their corresponding manual annotations from 3 LLMs as PRE's qualification exam data. Moreover, we select 7 LLMs as candidate LLMs. Among them, ChatGLM2-6B has evolved to ChatGLM3-6B during our work, so we choose ChatGLM3-6B as a candidate LLM while reusing the manual annotations of ChatGLM2-6B.

For DailyDialog, we uniformly sample 100 conversations from the original datasets to construct the question set. The average number of turns in these conversations is 7.67, with an average length of 99.15 words. Subsequently, we retain the candidate LLMs unchanged while selecting another 7 LLMs to generate the next sentence based on the conversation history. The average length of the generated responses is 16.14 words. After obtaining the question and answer sets, for each question-answer pair, we employ three human annotators to provide scores ranging from 1 to 5. Finally, Cohen’s Kappa value for the annotations is 0.8571, indicating a high level of inter-rater reliability. We take the median of these scores as the final annotation.

\begin{table}[t]
\centering
\setlength{\tabcolsep}{1mm} 
\begin{tabular}{ccccc}
\hline
LLM    & \begin{tabular}[c]{@{}c@{}}manual \\ annotation\end{tabular} & \begin{tabular}[c]{@{}c@{}}PRE \\ qualification \\ exam data\end{tabular} & \begin{tabular}[c]{@{}c@{}}candidate \\ LLM\end{tabular} \\ \hline
GPT-4                                                        &                                                              &                                                                           & \checkmark                                \\ \hline
Claude-1                                                         & \checkmark                                    &                                                                           &                                                          \\ \hline
GPT-3.5-turbo    & \checkmark                                    & \checkmark                                                 & \checkmark                                \\ \hline
Vicuna-7b-v1   & \checkmark                                    &                                                                           & \checkmark                                \\ \hline
ChatGLM2(3)-6B    & \checkmark                                    &                                                                           & \checkmark                                \\ \hline
RWKV-4-Raven-7B  & \checkmark                                    &                                                                           &                                                          \\ \hline
Alpaca-7b                                                    & \checkmark                                    & \checkmark                                                 &                                                          \\ \hline
FastChat-t5-3b   & \checkmark                                    & \checkmark                                                 & \checkmark                                \\ \hline
ChatGLM-Pro                                                 &                                                              &                                                                           & \checkmark                                \\ \hline
Baichuan2-13b                                               &                                                              &                                                                           & \checkmark                                \\ \hline
\end{tabular}

\caption{The basic information and all the uses of LLMs.}
\label{tab:4-1}
\end{table}
\begin{table}[t]
\small
\setlength{\tabcolsep}{1mm} 
\begin{tabular}{cl}
\hline
\begin{tabular}[c]{@{}c@{}}Evaluation \\ Format\end{tabular} & \multicolumn{1}{c}{Template}                                                                                                                                                                                                                                                                                                                                                                                                                                                                                                                    \\ \hline
5-level                                                      & \begin{tabular}[c]{@{}l@{}}\#\#\#Task Description\#\#\#.\\ Directly output a number between 1 and 5 \\ to indicate the quality score of this answer:\\ - 1 means the answer is irrelevant to the question\\ - 2 means the answer is related to the question,\\ but does not solve the question\\ - 3 means the answer only solves a part of the question\\ - 4 means the answer solves majority aspects of \\ the question, but not perfect \\ - 5 means the answer is perfect to solve the question \\ \#\#\#Question+Answer\#\#\#\end{tabular} \\ \hline
100-level                                                    & \begin{tabular}[c]{@{}l@{}}\#\#\#Task Description\#\#\#\\ Directly output a number between 0 and 100 to \\ indicate the score of this answer. The higher the score, \\ the higher the quality of the answer.\\ \#\#\#Question+Answer\#\#\#\end{tabular}                                                                                                                                                                                                                                                                                          \\ \hline
pairwise                                                     & \begin{tabular}[c]{@{}l@{}}\#\#\#Task Description\#\#\#\\ You only need to output `one' or `two' directly to \\ indicate which answer is better.\\ \#\#\#Question+Two Answers\#\#\#\end{tabular}                                                                                                                                                                                                                                                                                                                                                 \\ \hline
\end{tabular}
\caption{Evaluation formats and templates used for pointwise and pairwise assessment of answer quality.}
\label{tab:2-1}
\end{table}
\begin{table*}[t]
\centering 
\begin{tabular}{@{}ccccccc@{}}
\toprule
$L_1$ &
  $L_2$ &
  \begin{tabular}[c]{@{}c@{}}ChatGLM\\ 3-6B\end{tabular} &
  \begin{tabular}[c]{@{}c@{}}FastChat\\ -t5-3b\end{tabular} &
  \begin{tabular}[c]{@{}c@{}}Baichuan2\\ -13b\end{tabular} &
  \begin{tabular}[c]{@{}c@{}}ChatGLM\\ -Pro\end{tabular} &
  \begin{tabular}[c]{@{}c@{}}GPT-3.5\\ -turbo\end{tabular} \\ \midrule
GPT-4        & \multirow{3}{*}{\begin{tabular}[c]{@{}c@{}}Baichuan2\\ -13b\end{tabular}}                                                  & 0.765 & 0.865 & 0.670 & 0.995 & 0.980 \\
Vicuna-7b-v1 &                                                                                 & 0.740 & 0.855 & 0.670 & 0.990 & 0.975 \\
ChatGLM2-6B  &                                                                                 & 0.730 & 0.855 & 0.670 & 0.975 & 0.970 \\ \midrule
GPT-4        & \multirow{3}{*}{\begin{tabular}[c]{@{}c@{}}GPT-3.5\\ -turbo\end{tabular}} & 0.770 & 0.865 & 0.630 & 0.990 & 0.955 \\
Vicuna-7b-v1 &                                                                                 & 0.700 & 0.860 & 0.640 & 0.970 & 0.945 \\
ChatGLM2-6B  &                                                                                 & 0.705 & 0.820 & 0.650 & 0.970 & 0.950 \\ \midrule
GPT-4        & \multirow{3}{*}{\begin{tabular}[c]{@{}c@{}}self\\ -generate\end{tabular}} & 0.740 & 0.835 & 0.670 & 0.995 & 0.960 \\
Vicuna-7b-v1 &                                                                                 & 0.715 & 0.800 & 0.670 & 0.985 & 0.935 \\
ChatGLM2-6B  &                                                                                 & 0.695 & 0.765 & 0.670 & 0.985 & 0.960 \\ \bottomrule
\end{tabular}%
\caption{The $P_p$ of different candidate LLMs.}
\label{tab:5-3}
\end{table*}

\begin{table}[t]
\centering
\resizebox{\columnwidth}{!}{
\begin{tabular}{@{}lcc@{}}
\toprule
candidate LLMs (uncertainty $\downarrow$)      & easy set (mean±std)      & hard set (mean±std)  \\ \midrule
prompt1 + ChatGLM3-6B & 0.6090±0.2027 & 0.6407±0.1960 \\
prompt2 + ChatGLM3-6B   & \textbf{0.5573±0.1932} & \textbf{0.5015±0.1671} \\
prompt1 + Baichuan2-13b & 0.3198±0.2288 & 0.4088±0.1776 \\
prompt2 + Baichuan2-13b & 0.3631±0.2137 & 0.4308±0.1881 \\
prompt1 + GPT-3.5-turbo & 0.2871±0.3008 & 0.3776±0.2844 \\
prompt2 + GPT-3.5-turbo & 0.2102±0.2444 & 0.3269±0.2516 \\ \bottomrule
\end{tabular}%
}
\caption{The uncertainty extracted by probability transformation, with lower values indicating higher self-confidence.}
\label{tab:5-2}
\end{table}

\begin{table}[t]
\centering
\resizebox{\columnwidth}{!}{
\begin{tabular}{@{}lcc@{}}
\toprule
candidate LLMs (level labels $\uparrow$)       & easy set (mean±std)      & hard set (mean±std)   \\ \midrule
doubtful\_ChatGLM-Pro   & 4.0824±0.5333 & 3.7474±0.4796 \\
null\_ChatGLM-Pro       & 3.9738±0.4020 & 3.8883±0.4293 \\
doubtful\_ChatGLM3-6B   & 4.0331±0.8913 & 3.8101±0.9322 \\
null\_ChatGLM3-6B       & \textbf{3.9206±1.0979} & \textbf{4.0964±1.0155} \\
doubtful\_GPT-3.5-turbo & 4.2050±0.5683 & 3.7450±0.4359 \\
null\_GPT-3.5-turbo     & 4.0500±0.2179 & 3.9600±0.2417 \\ \bottomrule
\end{tabular}%
}
\caption{The self-confidence levels extracted by direct prompting, with higher values indicating higher self-confidence.}
\label{tab:5-1}
\end{table}

\subsection{2. Evaluation Prompts}
We compare two evaluation formats: pointwise and pairwise. For pointwise, we design two implementation approaches: 5-level and 100-level. For pairwise evaluation, we minimize the bias introduced by the positioning of answers by calculating the mean of the evaluation results before and after swapping the positions of the two answers. The specific prompts can be found in Table \ref{tab:2-1}.

\subsection{3. Implementation and Validation}
In this section, we validate the effectiveness of our proposed methods and introduce the implementation details.

\subsubsection{Consistency:} While PRE (Auto-Exam) has preliminarily validated the effectiveness of this method, we argue that its design and implementation appear overly simplistic. For example, the threshold value is rigidly fixed at 0.55, which lacks explainability and generalizability. To address this issue, we explore some dynamic approaches to threshold setting. Experimental results indicate that using the mean $P_c$ of all candidate LLMs as the threshold is an efficient, cost-effective, and adaptable solution. Accordingly, we adopt this method in our experiments. It will be interesting to explore other threshold-setting strategies for future research.

\subsubsection{Pertinence:} We conduct an experiment to validate the method's effectiveness. Specifically, we use Xsum as the question set $Q$ and employ GPT-4 to modify $Q$ into $Q'$. Then we explore the impact of LLMs that generate RA ($L_1$) and IA ($L_2$), selecting ChatGLM3-6B, FastChat-t5-3b, Baichuan2-13b, ChatGLM-Pro, and GPT-3.5-turbo as candidate LLMs. Table \ref{tab:5-3} shows that the candidate's $P_p$ generally decreases as $L_1$ weakens and $L_2$ strengthens. This trend can be attributed to weaker $L_1$ degrading RA quality and stronger $L_2$ enhancing IA, which in turn makes candidates more prone to being misled by IA. Notably, when $L_2$ is the candidate LLM itself (`self-generate'), the candidate's $P_p$ does not significantly differ from when $L_2$ is another LLM, suggesting the feasibility of self-generated IA. In our experiments, we use GPT-4 to modify $Q$ into $Q'$, set the threshold $\eta_p$ as the mean $P_p$ of all candidate LLMs, and designate $L1$ as ChatGLM2-6B and $L2$ as GPT-3.5-turbo.

\begin{figure}[t]
    \centering
    \includegraphics[width=0.9\columnwidth]{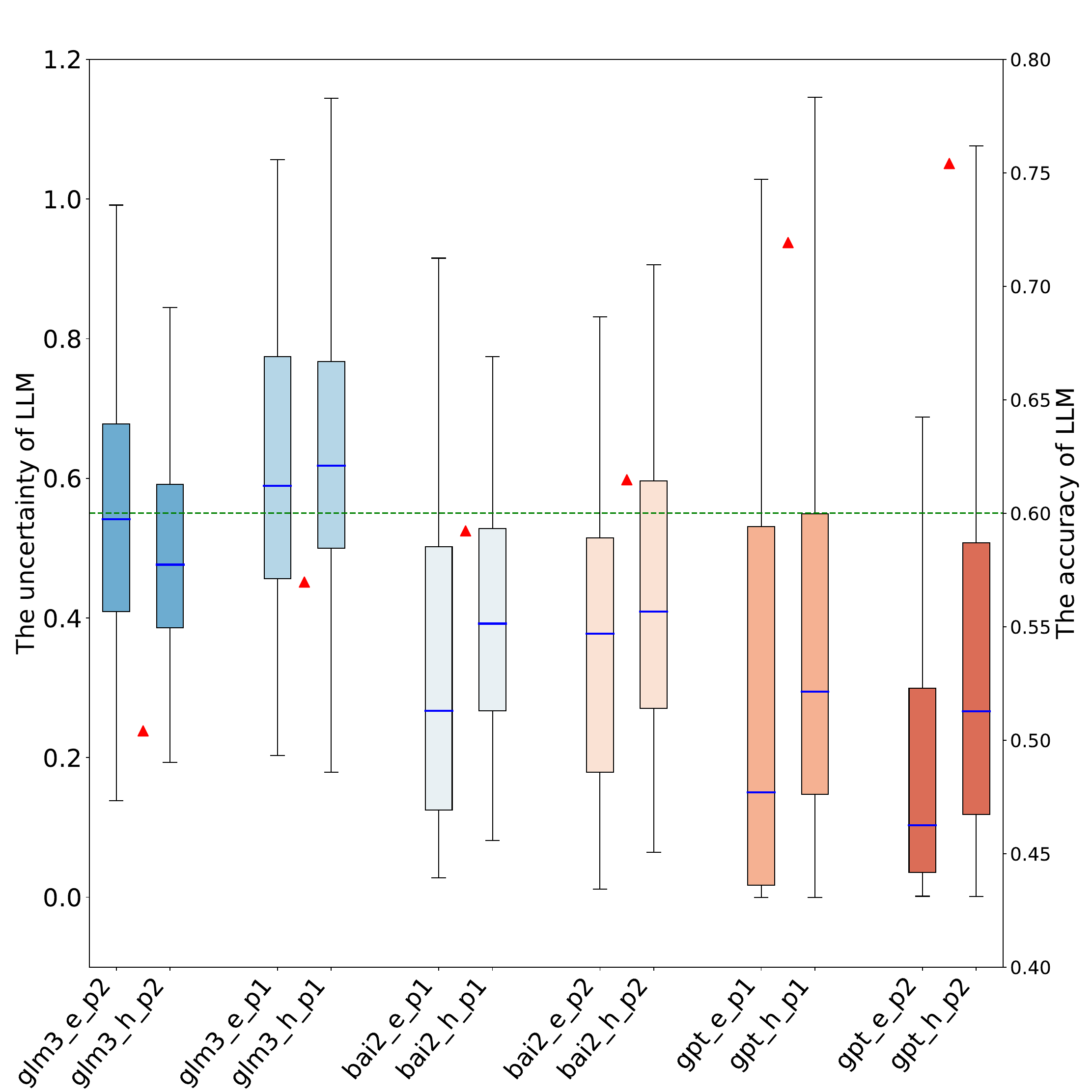}
    \caption{The left vertical axis represents LLM uncertainty. The right vertical axis shows LLM accuracy in manual annotation-based qualification exams. The horizontal axis displays experimental groups, divided into easy and hard sets (e.g., glm3\_e\_p2 denotes ChatGLM3-6B with prompt2 on the easy set). Accuracy is marked by red triangles, while uncertainty is illustrated using box plots \cite{williamson1989box}. A paired box with a higher median (blue line) on the left than on the right indicates unreasonable confidence levels.}
    \label{fig:5-1}
\end{figure}

\subsubsection{Self-Confidence:}  We present methods for extracting LLMs' self-confidence using probability transformation and direct prompting. For probability transformation, we constrain the model's output with the statement: "You only need to output one or two to indicate which answer is better." We design two prompts based on the statement's position: `\textit{prompt1}' places it at the beginning, and `\textit{prompt2}' places it at the end. For direct prompting, we design two strategies with different self-confidence level labels: [`doubtful', `uncertain', `moderate', `confident', `absolute'] (referred to as `\textit{doubtful}') and [`null', `low',  `medium', `high', `expert'] (referred to as `\textit{null}'). After obtaining self-confidence level labels, we convert them into integers from 1 to 5 (with higher numbers indicating greater confidence). Next, we explore whether our method can effectively filter out LLMs with unreasonable self-confidence levels. Table \ref{tab:5-2} and \ref{tab:5-1} show the self-confidence levels extracted by two methods on the Xsum dataset. The easy set is formed by GPT-4 and RWKV, and the hard set is formed by GPT-4 and Claude. Both results show that ChatGLM3-6B has lower self-confidence on the easy set than on the hard set, indicating unreasonable confidence, and is therefore filtered out. Then, we compare our filtering results with those of PRE (Human\_Filter), as shown in Figure \ref{fig:5-1}. The results indicate that LLMs with unreasonable confidence levels also exhibit lower accuracy in qualification tests based on manual annotations, which demonstrates the effectiveness of our method. In our experiments, we use GPT-4 and Claude to construct the hard set, GPT-4 and RWKV to construct the easy set, and adopt the `\textit{null}' and `\textit{prompt1}'.

\subsection{4. Detailed Settings Of Various Variants}
\begin{figure}[t]
    \centering
    \includegraphics[width=\columnwidth]{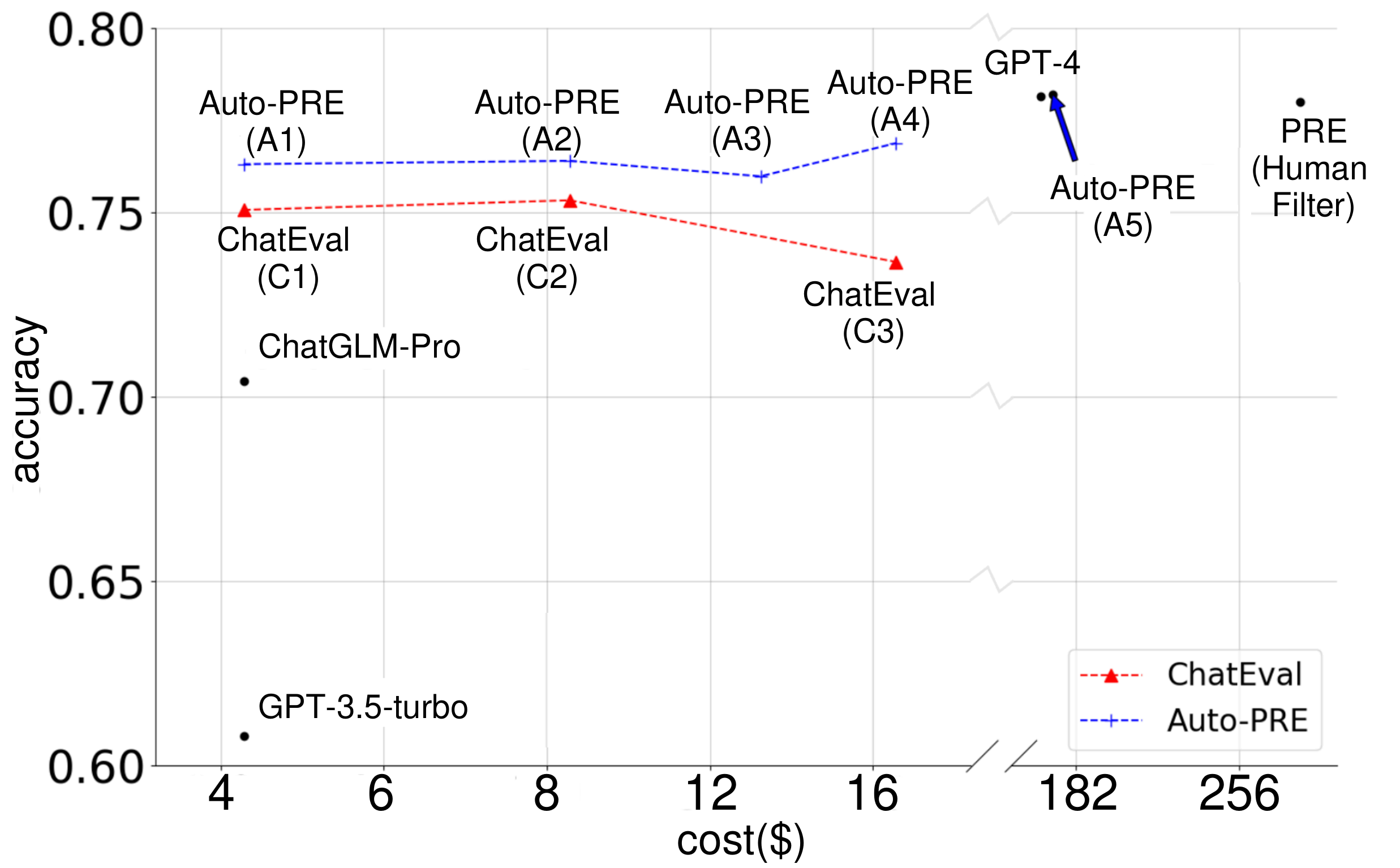}
    \caption{The performance on the NF\_CATS (pairwise).}
    \label{fig:5-3}
\end{figure}

\begin{table}[t]
\centering
\resizebox{\columnwidth}{!}{
\begin{tabular}{@{}ccc@{}}
\toprule
Methods &
  Settings &
  \begin{tabular}[c]{@{}c@{}}Cost\\(\$/M tokens) \end{tabular}
  \\ \midrule
ChatEval (C1) &
  \begin{tabular}[c]{@{}c@{}}one GPT-3.5-turbo as evaluator;\\ one role; one round of debate;\\ one-by-one strategy\end{tabular} &
  1 \\ \midrule
ChatEval (C2) &
  \begin{tabular}[c]{@{}c@{}}two GPT-3.5-turbo as evaluators;\\ two roles; one round of debate;\\ one-by-one strategy\end{tabular} &
  2 \\ \midrule
ChatEval (C3) &
  \begin{tabular}[c]{@{}c@{}}two GPT-3.5-turbo as evaluators, \\ two roles; two rounds of debate;\\ one-by-one strategy\end{tabular} &
  4 \\ \midrule
Auto-PRE (A1) &
  \begin{tabular}[c]{@{}c@{}}open-source LLMs; \\one ChatGLM-Pro\end{tabular} &
  1 \\ \midrule
Auto-PRE (A2) &
  \begin{tabular}[c]{@{}c@{}}open-source LLMs; \\one ChatGLM-Pro;\\ one GPT-3.5-turbo\end{tabular} &
  2 \\ \midrule
Auto-PRE (A3) &
  \begin{tabular}[c]{@{}c@{}}open-source LLMs; \\one ChatGLM-Pro;\\ two GPT-3.5-turbo\end{tabular} &
  3 \\ \midrule
Auto-PRE (A4) &
  \begin{tabular}[c]{@{}c@{}}open-source LLMs; \\two ChatGLM-Pro;\\ two GPT-3.5-turbo\end{tabular} &
  4 \\ \midrule
Auto-PRE (A5) &
  \begin{tabular}[c]{@{}c@{}}open-source LLMs; \\one ChatGLM-Pro;\\ one GPT-3.5-turbo; one GPT-4 \end{tabular} &
  42 \\
  \bottomrule
\end{tabular}%
}
\caption{The detailed settings and cost of various variants of evaluation methods. In Auto-PRE, if two same LLMs are used, the difference is the prompt strategy.}
\label{tab17}
\end{table}

In this section, we analyze the cost-effectiveness of Auto-PRE and compare its performance with ChatEval under equivalent costs. To achieve this, we implement several variants of the above methods. ChatEval includes three variants (C1, C2, C3), while Auto-PRE includes five variants (A1, A2, A3, A4, A5). Their detailed configurations are provided in Table \ref{tab17}. We use pairwise as the evaluation format on Xsum and NF\_CATS. Each task has 4200 instances, and each instance has about 1K tokens, so completing each task requires approximately 4.2 M tokens. Based on the official pricing released \cite{glmPricing, gpt3Pricing}, the costs of ChatGLM\_Pro and GPT-3.5-turbo are estimated to be similar at \$1 per million tokens. The cost of GPT-4 is estimated at \$40 per million tokens. Open-source LLMs are considered cost-free. Additionally, the cost of the qualification exam of PRE based on human annotations is about \$115 while the cost of our automatic qualification exam (less than \$1) can be neglected compared to the total costs.

\subsection{5. Hyperparameter Settings}

The PRE and its variants include two other important hyperparameters: threshold and weight: 

The threshold serves as the pass line for candidate LLMs in the exam. Following the original PRE paper, we set the threshold to 0.6 for PRE (Human Filter) and 0.55 for PRE (Auto-Exam). 

The weight refers to the fusion weight of each evaluator LLM when merging all evaluation results in the final stage. Unless otherwise specified, the weight is set to 1 by default in our experiments. For PRE (Human Filter), we follow the original PRE and assign the weight based on the score achieved in the qualification exam.

In addition to the above, unless otherwise specified, the results reported in our experiments represent the best performance achieved by Auto-PRE, using a combination of three selection methods based on task requirements. The candidate LLMs are identical to those used in PRE (Human Filter). The weight of Auto-PRE is calculated as the average of the scores from the three selection methods ($P_c$, $P_p$, and $P_s$).
\end{document}